\documentclass[letterpaper, 10 pt, conference]{ieeeconf}  % Comment this line out if you need a4paper
\IEEEoverridecommandlockouts
\usepackage{amsmath,amssymb,xcolor}
\usepackage[]{graphicx}

\usepackage{algpseudocode}
\usepackage{algorithm}
\usepackage{hyperref}

\newcommand{\vb}{\mathbf{b}}

\newcommand{\vd}{\mathbf{d}}

\newcommand{\vu}{\mathbf{u}}
\newcommand{\vv}{\mathbf{v}}

\newcommand{\vx}{\mathbf{x}}
\newcommand{\vy}{\mathbf{y}}
\newcommand{\vz}{\mathbf{z}}

\newcommand{\cL}{\mathcal{L}}
\newcommand{\vo}{\mathbf{0}}

\newcommand{\vA}{\bf{A}}
\newcommand{\vL}{\bf{L}}
\newcommand{\st}{\mbox{s.t.}}
\DeclareMathOperator*{\sign}{sign}

\DeclareMathOperator{\prox}{prox}
\DeclareMathOperator{\shrink}{shrink}

\DeclareMathOperator*{\argmin}{argmin}
\newcommand{\R}{\mathbb{R}}

\newcommand{\norm}[1]{\lVert#1\rVert}
\graphicspath{{./figs/}}

\title{\huge Hand Gesture Recognition Based on a Nonconvex Regularization}

\author{\authorblockN{Jing Qin}
\authorblockA{\textit{Department of Mathematics}\\
\textit{University of Kentucky}\\
\textit{Lexington, KY 40506, USA}\\
\textit{jing.qin@uky.edu}\\}%
\and
\authorblockN{Joshua Ashley and Biyun Xie}
\authorblockA{\textit{Department of Electrical and Computer Engineering}\\
\textit{University of Kentucky}\\
\textit{Lexington, KY 40506, USA}\\
\textit{\{jaas224,biyun.xie\}@uky.edu}\\}}%

\begin{document}

\maketitle
%\thispagestyle{empty}
%\pagestyle{empty}

%%%%%%%%%%%%%%%%%%%%%%%%%%%%%%%%%%%%%%%%%%%%%%%%%%%%%%%%%%%%%%%%%%%%%%%%%%%%%%%%
\begin{abstract}
Recognition of hand gestures is one of the most fundamental tasks in human-robot interaction. Sparse representation based methods have been widely used due to their efficiency and low demands on the training data. Recently, nonconvex regularization techniques including the $\ell_{1-2}$ regularization have been proposed in the image processing community to promote sparsity while achieving efficient performance. In this paper, we propose a vision-based hand gesture recognition model based on the $\ell_{1-2}$ regularization, which is solved by the alternating direction method of multipliers (ADMM). Numerical experiments on binary and gray-scale data sets have demonstrated the effectiveness of this method in identifying hand gestures.

\end{abstract}


%%%%%%%%%%%%%%%%%%%%%%%%%%%%%%%%%%%%%%%%%%%%%%%%%%%%%%%%%%%%%%%%%%%%%%%%%%%%%%%%
\section{Introduction}
Human-robot interaction has become a popular research topic which can be integrated into and revolutionize almost every aspect of our lives.
Similar to human-human interaction, there are many ways for humans to express their intentions or emotions in human-robot interaction, which can be classified into two categories: verbal and nonverbal communications \cite{aly2013model}. Nonverbal communication further includes facial expressions \cite{wu2019weight}, gestures \cite{chang2019improved}, proxemics \cite{patompak2019learning}, and eye gazes \cite{saran2018human}. Verbal communication has the advantage of simplicity, convenience, and clearness. Nonverbal communication, however, is an essential interaction way for scenarios where verbal communication is not available, such as noisy environments and long-range interaction \cite{barattini2012proposed}. Even when verbal communication is available, nonverbal communication can also be used as a considerable augmentation of verbal communication, which will make the interaction more lively. A socially intelligent robot typically has the capacity of understanding human intentions through nonverbal communication to improve the effectiveness, efficiency, and human-friendless in human-robot interaction. In particular, hand gestures serve as a natural and intuitive way to assist interaction between humans and robots. Therefore, recognition of hand gestures plays a key role in a variety of human-robot interaction applications.

There are mainly two types of hand gesture recognition. For the first type, glove-based hand gesture recognition, hand gestures are recorded using a data glove, and the position of each finger joint can be obtained accordingly. For the other type, vision-based hand gesture recognition, hand gestures are captured using cameras. Compared with data gloves, camera-based capture systems are much cheaper and easier to use. Besides, wearing gloves will cause difficulty for some hand operations, such as clenching fists. In this paper, we will focus on vision-based hand gesture recognition.

In pattern recognition, sparse representation has shown its great power in compressing and processing high-dimensional data. Specifically, we assume that the object to be recognized can be sparsely represented as a linear combination of atoms in a redundant dictionary, which implies a large portion of coefficients are zeros. In order to find this sparse representation, we can resort to sparsity based regularization techniques. For example, $\ell_1$-regularization has been applied to dictionary-based action recognition \cite{qiu2011sparse} and its local version has been proposed for gesture recognition \cite{he2019gesture}. Sparsity-based methods typically improve interpretability and compressibility of data, which enable detection discriminative and avoid over-fitting. Recently, the nonconvex $\ell_{1-2}$ regularization has shown to promote higher sparsity and achieve better performance than its $\ell_1$ counterpart in image reconstruction \cite{yin2015minimization,li2016s} and in logistic regression \cite{qin20191}. In light of this, we propose a novel $\ell_{1-2}$-regularized hand gesture recognition model, which is then solved by applying the alternating direction method of multipliers (ADMM). Each resultant subproblem has a closed-form solution which leads to computational efficiency. Note that the proposed sparsity-based model is not a trivial generalization of that in \cite{miao2016gesture}, which considers an inequality constrained $\ell_1$-minimization problem different from our proposed model. In addition, we will take advantage of various features, including binary segmented images, histograms of oriented gradients (HOG) \cite{MERL_TR94-03} and local binary patterns (LBP) \cite{ojala1996comparative}. HOG uses the distribution of intensity gradients along various orientations to describe local object appearance and shape within an image. By contrast, LBP exploits local binary patterns over an image. In pattern recognition, HOG and LBP have been shown to be effective and robust feature descriptors for object detection \cite{ghorbani2015hog}. To verify the effectiveness of the proposed method, we test two sets of hand gesture images in binary or gray scales. Performance of the method in terms of recognition rate and running time under various settings of training samples are reported. We also make discussions on parameter selection, cell size in HOG and LBP, identification metric, and comparison between $\ell_1$ and $\ell_{1-2}$.

The rest of this paper is organized as follows. In Section \ref{sec:SR}, we provide a brief introduction of sparse representation based models. In Section \ref{sec:method}, we propose a novel hand gesture recognition algorithm based on the nonconvex $\ell_{1-2}$ regularization. Numerical experiments on two realistic data sets of hand gestures and the results are discussed in Section \ref{sec:exp}. Finally, conclusions of this research and future work are presented in Section \ref{sec:con}.

\section{Sparse Representation Based Models}\label{sec:SR}
Throughout the paper, we use boldface lowercase letters to denote vectors and boldface uppercase letters to denote matrices. For $p\in\mathbb{N}$, the $\ell_p$-norm of a vector $\vx\in\R^n$ is defined as $\norm{\vx}_p=(\sum_{i=1}^n|x_i|^p)^{1/p}$.

Assume that a test vector $\vb\in\R^{m}$ can be sparsely represented as a linear combination of columns in a dictionary $\Phi$, i.e., there exists $\vx\in\R^n$ with small $\norm{\vx}_0$ such that $\vb=\Phi\vx$. Here $\norm{\vx}_0$ is the number of nonzero components in $\vx$ and can describe the sparsity of the vector $\vx$. Since the dictionary is redundant, the image size is much smaller than the number of images in the dictionary, i.e., $m\ll n$, which results in infinitely many solutions to the linear system $\Phi\vx=\vb$. To guarantee a unique solution, we consider the $\ell_0$ minimization problem of the form
\[
\min_{\vx\in\R^n}\norm{\vx}_0\quad\st\quad \Phi\vx=\vb.
\]
However, this problem is NP-hard which can be relaxed to the convex $\ell_1$ minimization
\[
\min_{\vx\in\R^n}\norm{\vx}_1\quad\st\quad \Phi\vx=\vb.
\]
To further enforce sparsity on the solution, we consider the $\ell_{1-2}$ minimization
\begin{equation}\label{eqn1}
\min_{\vx\in\R^n}\norm{\vx}_{1-2}\quad \st\quad \Phi\vx=\vb,
\end{equation}
where $\norm{\vx}_{1-2}=\norm{\vx}_1-\beta\norm{\vx}_2$. It has empirically shown that the choice of $\beta$ does not make significant impact on the performance. Thus we fix $\beta=1$ to reduce the number of tuning parameters throughout the paper. Note that $\ell_{1-2}$ is not a vector norm in $\R^n$ since the triangle inequality and positive definiteness cannot be guaranteed. Connections and comparisons between the $\ell_{1-2}$ regularization and its $\ell_1$ counterpart can be referred to \cite{li2016s,qin20191}.

\section{Proposed Method}\label{sec:method}
Recognition of hand gestures plays an important role in the human-robot interaction. In particular, vision-based recognition methods aim to identify the gesture pattern from a dictionary (also known as library) of images that is most similar to the test image. Each hand image in the dictionary is called an \emph{atom}.

\subsection{Dictionary Construction}
There are many types of dictionaries that can be used for gesture recognition, where each atom can characterize one or multiple features of an image. One simple example is to use binary segmented images as atom images which separates the hand from the background. However, once we reshape each image as a column vector, a sparse representation of atoms may not be sufficient to describe the image geometric information. To further take local geometry into consideration, we can create a dictionary of HOG or LBP features (see Section~\ref{sec:exp} for more details).

In the continuous setting, each image can be considered as a function $f:\Omega\to\mathbb{R}$ where $\Omega\subseteq\mathbb{R}^2$ is a closed set with Dirichlet type of boundaries, e.g., $\Omega=[a,b]\times[c,d]$. Suppose that $f$ can be sparsely represented by the set of atoms $\{v_1,\ldots,v_n\}$ where each atom image $v_i:\Omega\to\R$, meaning that coefficients $c_1,\ldots,c_n$ exist with $f(x,y)=\sum_{i=1}^nc_iv_i(x,y)$
and the number of nonzero coefficients $c_i$'s is small. If we restrict the domain of this function on a grid, then sparse representation of a discrete image in terms of discrete atoms still holds locally which implies that sparse representation in the feature space can still be preserved. In this work, we adopt three types of dictionaries using binary or gray-scale segmented images, HOG and LBP features.

\subsection{$\ell_{1-2}$-Regularized Recognition Method}
Starting from this section, we will consider discrete images, i.e., each image is treated as a matrix. Given $n$ atoms of size $\sqrt{m}\times\sqrt{m}$, we reshape each image as a column vector by columnwise stacking and then concatenate them to form a dictionary $D=[\vd_1,\ldots,\vd_n]\in\mathbb{R}^{m\times n}$. Similarly, the test image $\vb$ is reshaped as a column vector.

Suppose there are $L$ classes in the dictionary $D$ corresponding to $L$ gestures, i.e., the dictionary $D$ can be partitioned as $D=\bigcup_{i=1}^L D_i$ each $D_i\in\R^{m\times n_i}$ such that $D_i$ and $D_j$ have disjoint columns for $i\neq j$ and $n_1+\ldots+n_L=n$. Without loss of generality, let $\vA$ be one such sub-dictionary $D_i$. If the partition is not available, we can apply fast data clustering algorithms such as $k$-means.

Next, we intend to find a sparse representation of the test data $\vb$ with respect to the dictionary $\vA$, i.e., finding $\vx$ with the smallest number of nonzero elements such that $\vA\vx=\vb$. First, we normalize the columns of $A$ so that every column has a unit $\ell_2$-norm. Then we consider the $\ell_{1-2}$-regularized sparse recovery model
\begin{equation}\label{eqn2}
\min_{\vx}\frac12\norm{\vA\vx-\vb}_2^2+\lambda\norm{\vx}_{1-2}.
\end{equation}
Here $\lambda>0$ is a regularization parameter. Different from the linear constrained model \eqref{eqn1}, the unconstrained model \eqref{eqn2} considers the presence of noise. By the change of variable, \eqref{eqn2} can be written as
\[
\min_{\vx,\vy}\frac12\norm{\vA\vx-\vb}_2^2+\lambda\norm{\vy}_{1-2}\quad\st\quad \vx=\vy.
\]
To solve this minimization problem, we define the augmented Lagrange function as follows
\[
\cL=\frac12\norm{\vA\vx-\vb}_2^2+\lambda\norm{\vy}_{1-2}+\frac{\rho}2\norm{\vx-\vy+\widehat{\vy}}_2^2.
\]
Here $\rho>0$ is a tuning parameter which controls the convergence speed. Following the framework of ADMM, we alternate the minimization of $\cL$ with respect to $\vx$ and $\vy$, respectively.
Notice that the subproblem $\argmin_{\vx}\cL$ is a least-square problem which can be converted to solving its normal equation. Hence we obtain the following updating scheme
\begin{equation}\label{eqn:admm}
\left\{
\begin{aligned}
\vx&\leftarrow (\vA^T\vA+\rho I)^{-1}(\vA^T\vb+\rho(\vy-\widehat{\vy}));\\%\vA^T(\vA\vx-\vb)+\rho(\vx-\vy+\widehat{\vy});\\
\vy&\leftarrow \prox_{\theta\norm{\cdot}_{1-2}}(\vx+\widehat{\vy});\\
\widehat{\vy}&\leftarrow \vx-\vy+\widehat{\vy},
\end{aligned}
\right.
\end{equation}
where $\theta=\frac{\lambda}{\rho}$. To accelerate the computation, we apply the Cholesky factorization of the matrix $\vA^T\vA+\rho I=\vL\vL^T$ with $\vL$ a lower triangular matrix and thereby the update of $\vx$ becomes
\begin{equation}\label{eqn:x_update}
\vx\leftarrow \vL^{-T}(\vL^{-1}(\vA^T\vb+\rho\vy-\rho\widehat{\vy})).
\end{equation}
where $\vL^{-1}$ is the inverse of the matrix $\vL$, and $\vL^{-T}$ is the transpose of $\vL^{-1}$.
Moreover, the proximal operator of a function $f$ is defined as $\prox_{f}(\vv)=\argmin_{\vu}\frac12\norm{\vu-\vv}_2^2+f(\vu)$. Note that the proximal operator of $\ell_{1-2}$ can be expressed as \cite{qin20191}:
\begin{equation}\label{eqn:y_update}
\prox_{\theta\norm{\cdot}_{1-2}}(\vx+\widehat{\vy})=\vz+\frac{\theta\vz}{\norm{\vz}_2},
\end{equation}
where $\vz=\shrink(\vx+\widehat{\vy},\theta)$ and the shrinkage operator is defined componentwise
\[
[\shrink(\vu,\mu)]_i=\sign(u_i)\max\{|u_i|-\mu,0\}
\]
for $i=1,2,\ldots,n$. Here $u_i$ is the $i$-th component of the vector $\vu$.
The algorithm terminates if either the relative change of two consecutive estimates of $\vx$ reaches the preassigned tolerance, i.e.,
\begin{equation}\label{eqn:stop}
\norm{\vx^{(j+1)}-\vx^{(j)}}_2/\norm{\vx^{(j)}}_2<tol
\end{equation}
where $\vx^{(j)}$ is the estimate of $\vx$ after $j$ iterations, or the maximal number of allowed iterations is achieved. From this step, we get the optimal coefficient vector $\vx^*$ of $\vb$ with respect to the dictionary $\vA$. Furthermore, we let $\vx_i^*$ be the solution to \eqref{eqn2} when $\vA=D_i$.

To identify the most similar gesture class, we adopt two types of identification metrics for classification. One metric uses the $\ell_2$-norm residual for each class given by
\begin{equation}\label{eqn:ri1}
r_i(\vb)=\norm{\vb- D_i\vx_i^*}_2,\quad i=1,\ldots,L,
\end{equation}
where $\vx_i^*$ is obtained in the previous set with $\vA=D_i$. Alternatively, we compare the cosine similarity between $\vb$ and $D_i\vx_i^*$ and define the identification metric as
\begin{equation}\label{eqn:ri2}
r_i(\vb)=1-\cos(\vb,D_i\vx_i^*),\quad i=1,\ldots,L.
\end{equation}
Here cosine similarity is defined as $\cos(\vu,\vv)=\frac{\norm{\vu}_2\cdot\norm{\vv}_2}{\langle \vu,\vv\rangle}$ where $\langle \cdot,\cdot\rangle$ is the dot product of two vectors.
Finally, we predict the class that $\vb$ belongs to by finding the minimum identification metric
\begin{equation}\label{eqn:id}
identity(\vb):=\argmin_{1\leq i\leq L}r_i(\vb).
\end{equation}
Other similarity metrics could be used to define $r_i(\vb)$ while it may take more computational time. The entire algorithm is summarized in Algorithm~\ref{alg}, which can be extended to recognize multiple test data points in parallel.

\begin{algorithm}
\caption{Nonconvex $\ell_{1-2}$-Regularized Recognition}\label{alg}
\begin{algorithmic}
\State\textbf{Inputs}: a dictionary with partition $D=\cup_{i=1}^LD_i$, test data $\vb$, parameters $\lambda,\rho>0$, maximum number of inner loops $N_{in}$, tolerance $tol$ for the stopping criterion
\State\textbf{Outputs}: class label of $\vb$
\For{$i=1,2,\ldots,L$}
    \State Initialize $\vx=\vo$
    \For{$j=1,2,\ldots,N_{in}$}
        \State Update $\vx$ via \eqref{eqn:x_update}
        \State Update $\vy=\prox_{\theta\norm{\cdot}_{1-2}}(\vx+\widehat{\vy})$ via \eqref{eqn:y_update}
        \State $\widehat{\vy}\leftarrow \vx-\vy+\widehat{\vy}$
        \State {Exit the inner loop if \eqref{eqn:stop} is met.}
    \EndFor
    \State $\vx_i^*=\vx$
    \State Calculate $r_i(\vb)$ via \eqref{eqn:ri1} or \eqref{eqn:ri2}.
\EndFor
\State Find the class label of $\vb$ via \eqref{eqn:id}.
\end{algorithmic}
\end{algorithm}

\section{Numerical Experiments}\label{sec:exp}
In this section, we will test the proposed Algorithm~\ref{alg} on one binary and one gray-scale data sets of hand gesture images. To quantify the performance, we use the recognition rate that is defined as the ratio of the correctly recognized labels out of the entire test set. To make comparison fair, we randomly select the test and atom images from the data set and take the average performance of 50 trials by default. There are three types of feature vectors being used: (1) raw feature vectors that are generated by reshaping each image as a vector via column-wise stacking; (2) reshaped HOG feature vectors with the cell size $k\times k$; (3) reshaped LBP feature vectors with the cell size $k\times k$. Both HOG and LBP extractions are implemented in Matlab. By default, the parameters of Algorithm~\ref{alg} are set as $\rho=1000,\lambda=1$. The maximum number of inner loops is set as 20 and the tolerance in \eqref{eqn:stop} is $tol=10^{-4}$. The cell size for both HOG and LBP is set as $k=8$. Note that even with the same cell size, HOG and LBP features do not have the same dimension. All experiments were run in Matlab R2019a on a desktop computer with Intel CPU i9-9960X RAM 64GB and GPU Dual Nvidia Quadro RTX5000 with Windows 10 Pro.

\subsection{Experiment 1: Binary Data}
\begin{figure}[h]
\centering\setlength{\tabcolsep}{1pt}
\begin{tabular}{ccc}
\includegraphics[width=.16\textwidth]{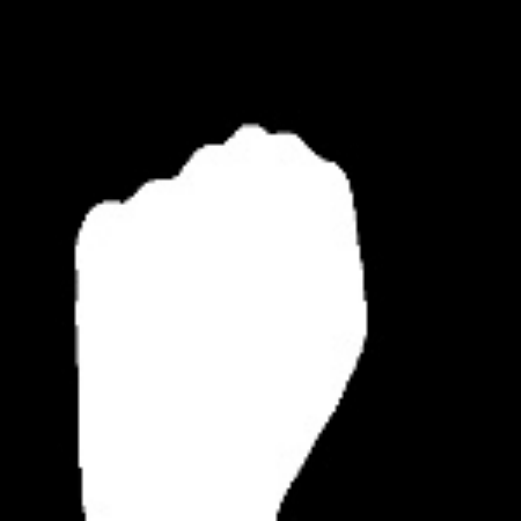}&
\includegraphics[width=.16\textwidth]{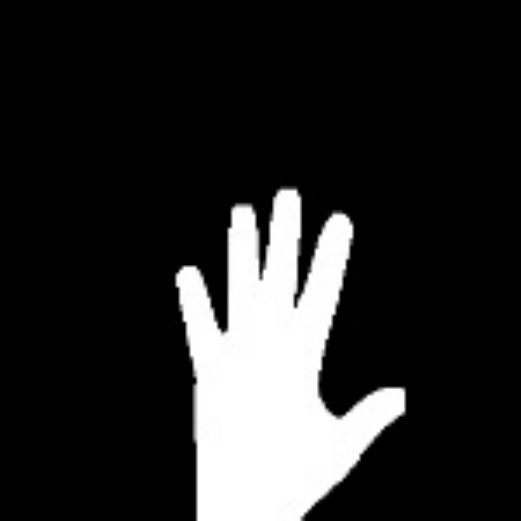}&
\includegraphics[width=.16\textwidth]{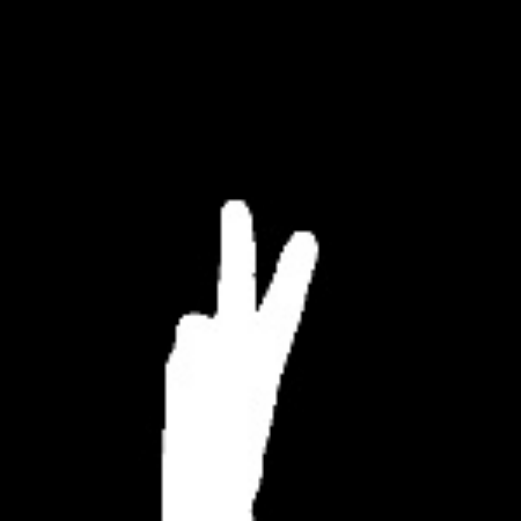}
\end{tabular}
\caption{Sample images in a binary dictionary. From left to right: fist, open hand, two fingers.}\label{fig1}
\vspace{-12pt}
\end{figure}

The first set of data is downloaded from \cite{shivamgoyal_2020}. Specifically, there are three hand gestures in the database: fist, open-hand, and two-finger, which have 2003, 2010 and 2005 images, respectively. Each image is binary of the size $150\times 150$. We select 10 images randomly from each gesture class and select $Nt$ images from the rest as the atoms in the test dictionary. Fig.~\ref{fig1} displays one sample image for each type of gestures.

\begin{figure}[H]
\centering
\includegraphics[width=.34\textwidth]{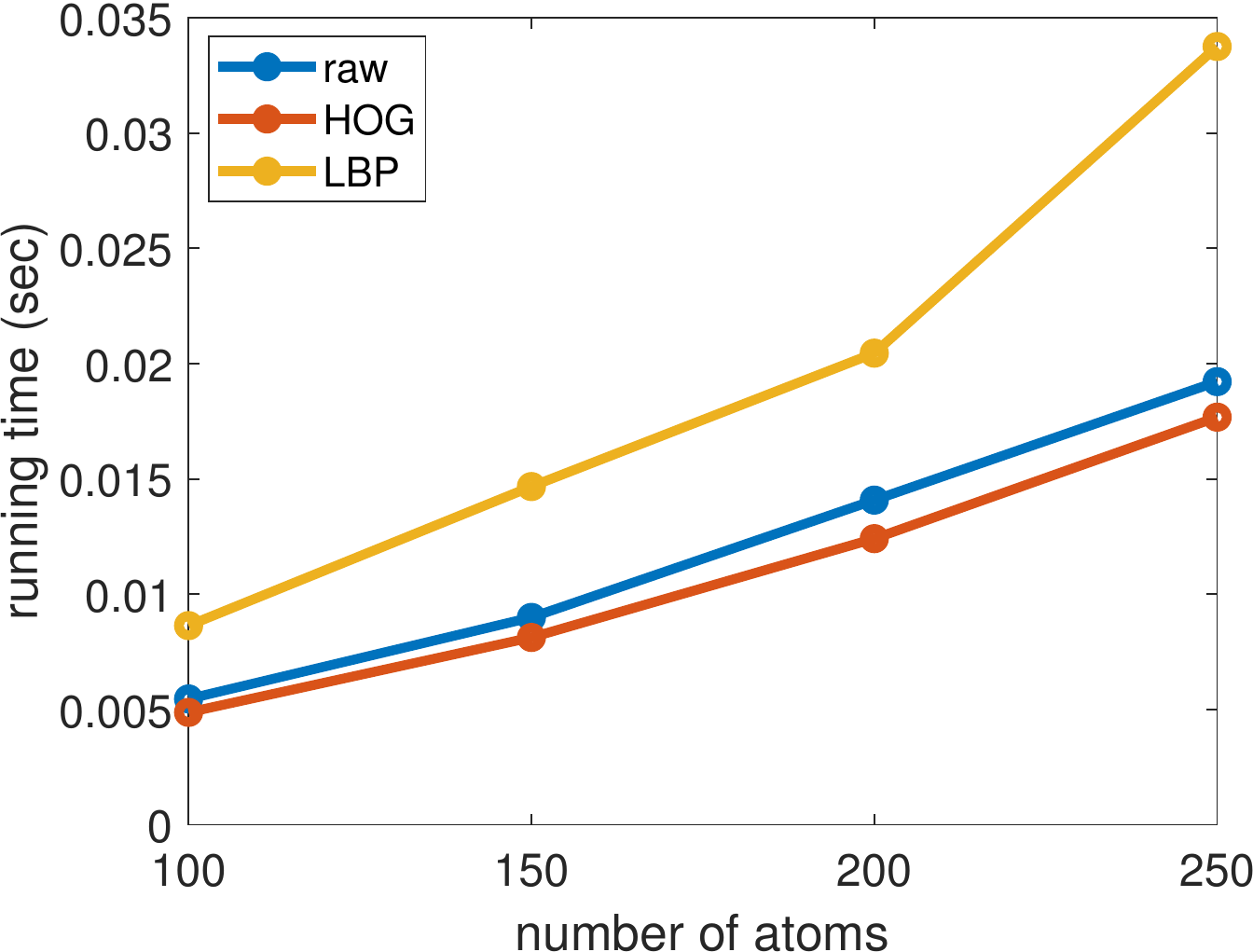}
\vspace{-8pt}
\caption{Running time comparison for a binary dictionary.}\label{fig:exp1_time}
\vspace{-10pt}
\end{figure}

We set the number of atoms in the test dictionary as $Nt=100, 150, 200, 250$, respectively. The recognition rates for all cases are shown in Table~\ref{tab1}. One can see that HOG type of features yields the best performance. In the meanwhile, since each image is piecewise constant with limited texture-like patterns, LBP performs the worst which agrees with the fact that LBP features favor the texture patterns \cite{alhindi2018comparing}. When the number of atoms is larger than 300, the proposed method can achieve almost perfect recognition. Comparison of average running time for each case is shown in Fig.~\ref{fig:exp1_time}. With the fixed cell size $8\times 8$, the dimension of each HOG feature is 10404 while 19116 for that of each LBP feature which explains why LBP spends the most running time.

\begin{table}[h]
\centering
\caption{Recognition rates on a binary dictionary.}\label{tab1}
\vspace{-6pt}
\begin{tabular}{ccccc}
\hline \hline
Feature $\backslash$ Atom No. & 100 & 150 & 200 & 250 \\ \hline
raw & 0.8060 & 0.8453 & 0.8800 & 0.8953 \\
HOG & 0.9080 & 0.9393 & 0.9520 & 0.9667 \\
LBP & 0.7373 & 0.7633 & 0.8053 & 0.8460 \\
\hline\hline
\end{tabular}
\vspace{-6pt}
\end{table}

\subsection{Experiment 2: Gray-Scale Data}\label{subsec:exp2}
\begin{figure*}[h]
\centering\setlength{\tabcolsep}{2pt}
\begin{tabular}{ccccc}
\includegraphics[width=0.19\textwidth]{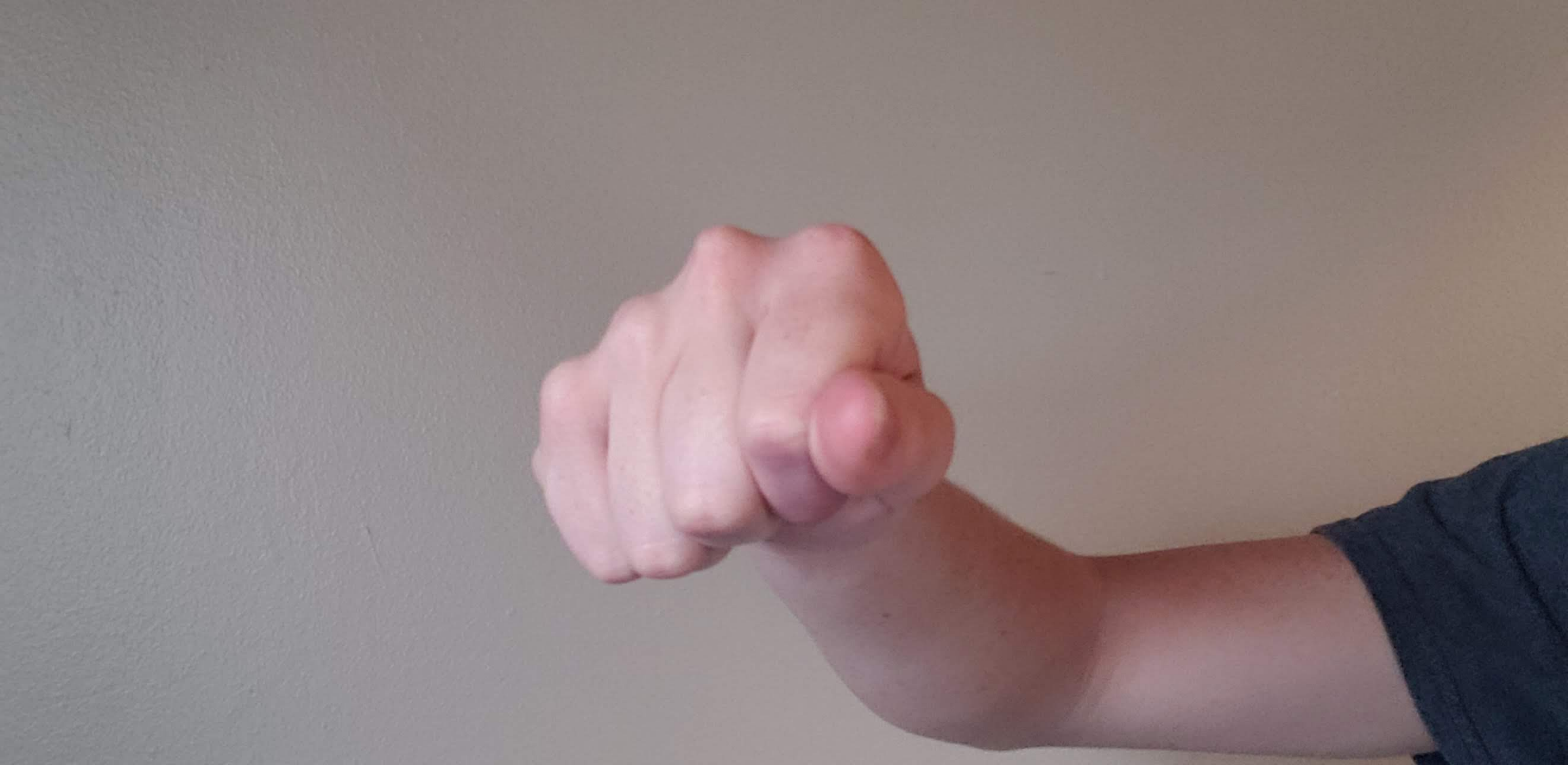}&
\includegraphics[width=0.19\textwidth]{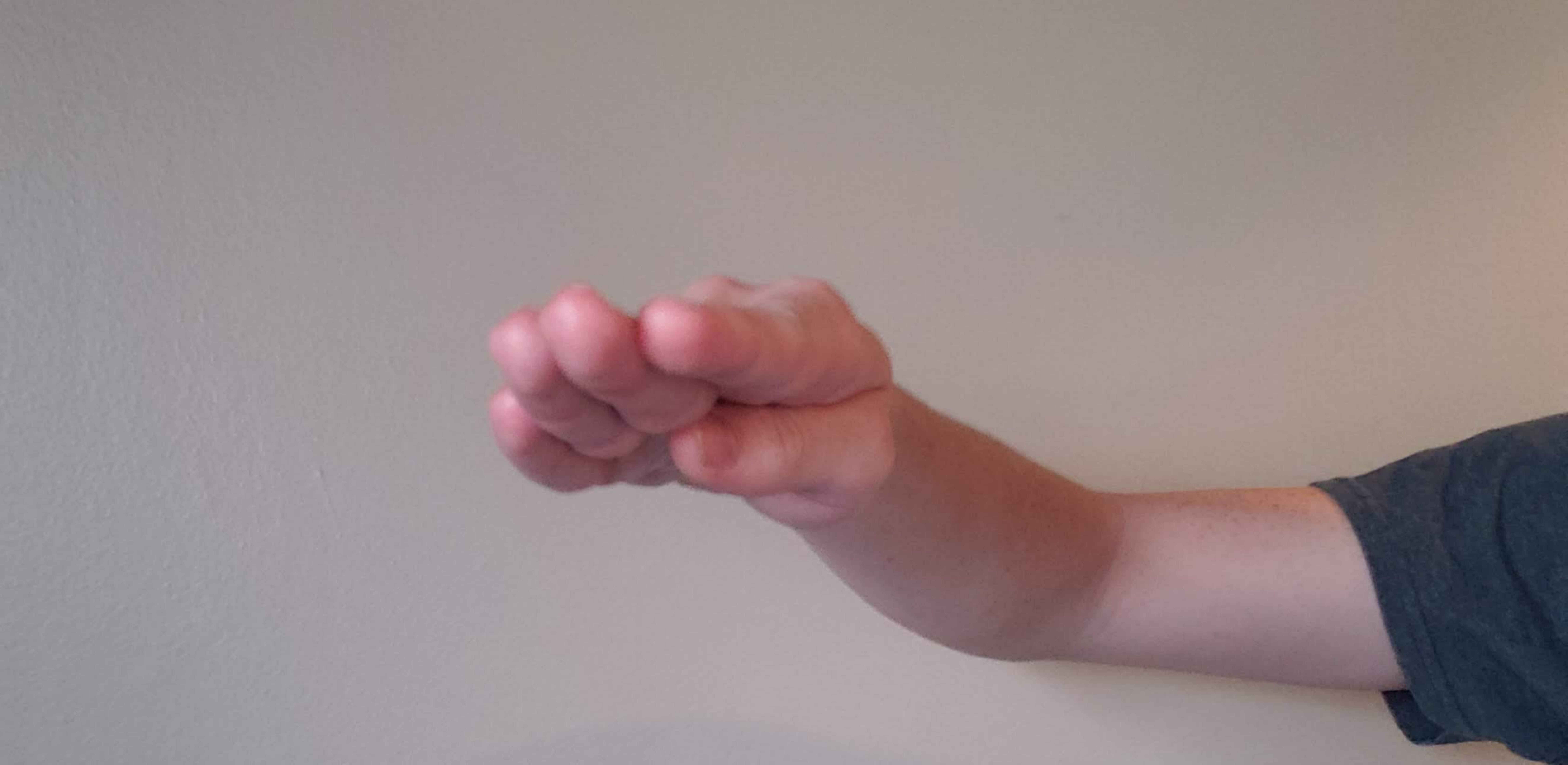}&
\includegraphics[width=0.19\textwidth]{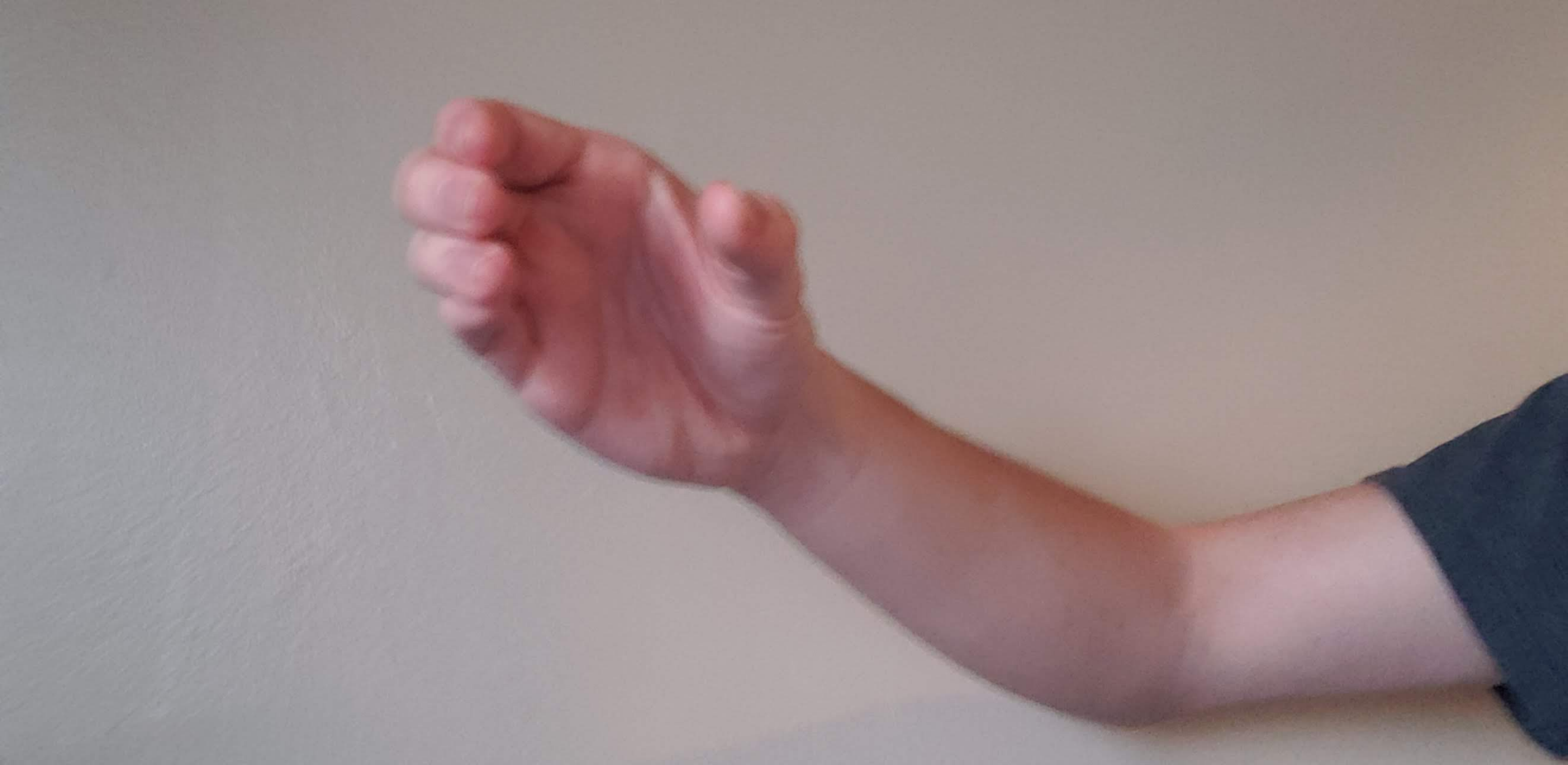}&
\includegraphics[width=0.19\textwidth]{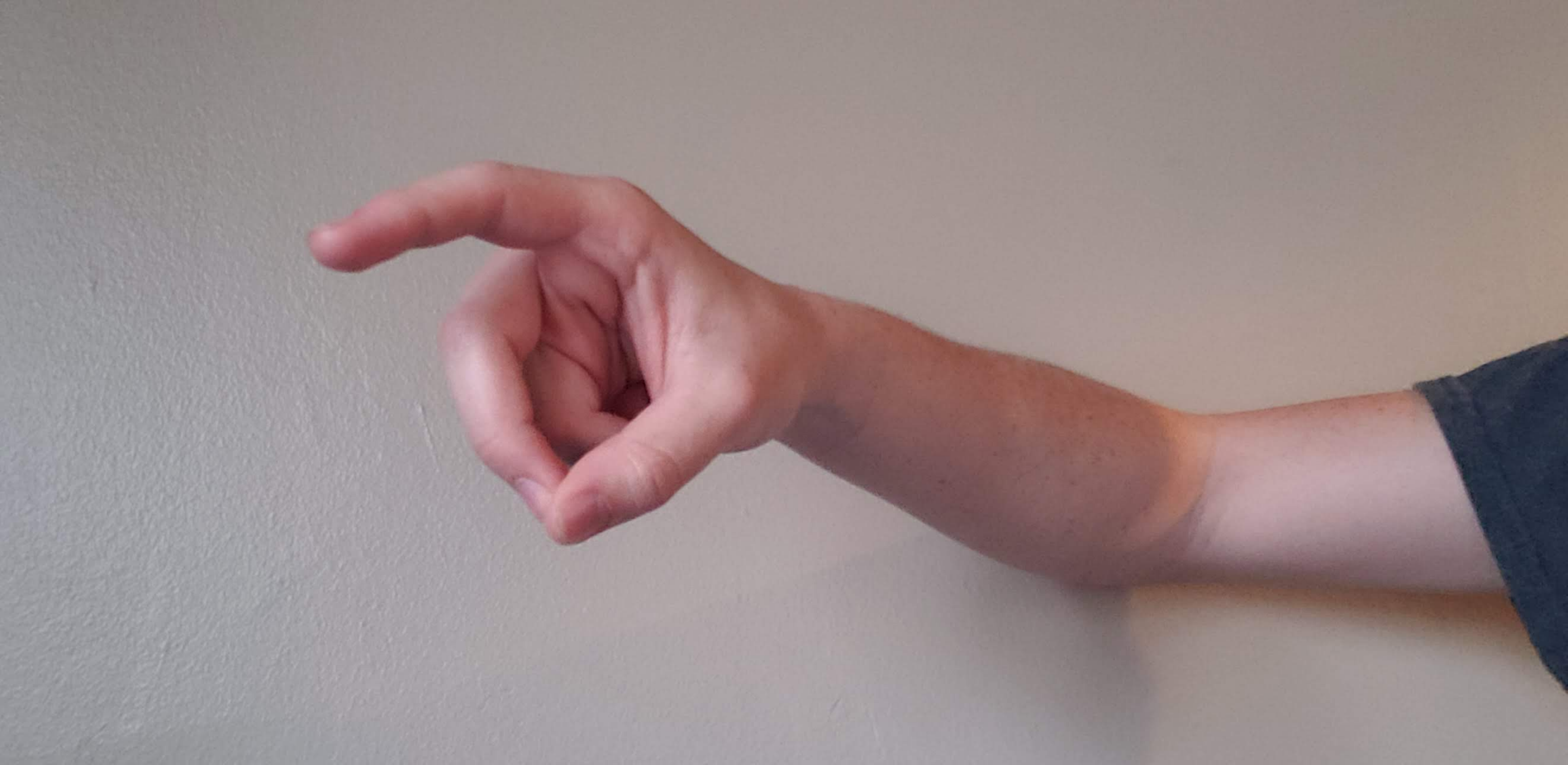}&
\includegraphics[width=0.19\textwidth]{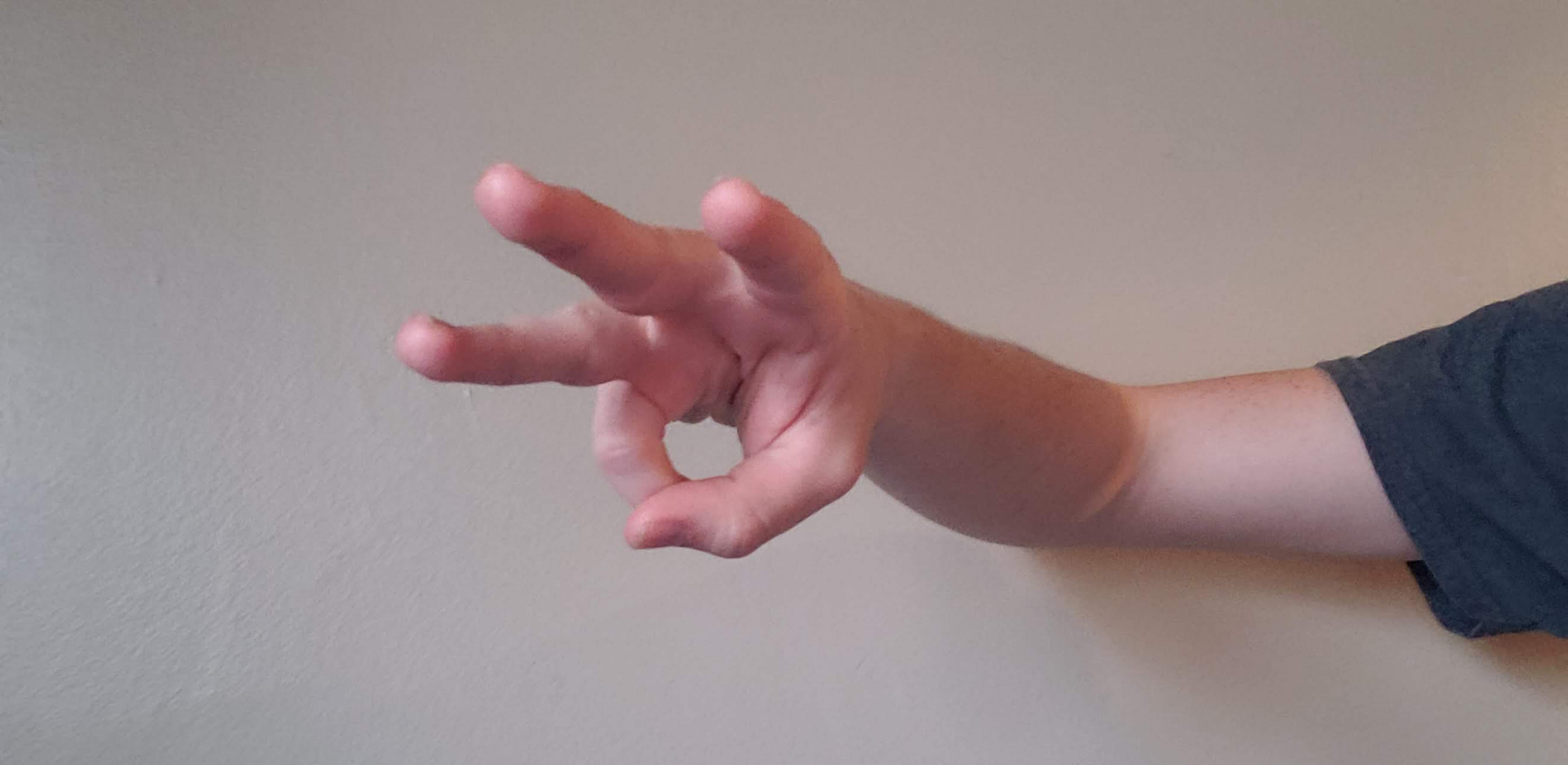}\\
(a) A & (b) B & (c) C & (d) D & (e) W
\end{tabular}
\vspace{-4pt}
\caption{Sample gesture images. The gestures from left to right correspond to the letters: A, B, C, D and W.}\label{fig2}
\end{figure*}

\begin{figure*}[h]
\centering\setlength{\tabcolsep}{2pt}
\begin{tabular}{ccccc}
\includegraphics[width=0.19\textwidth]{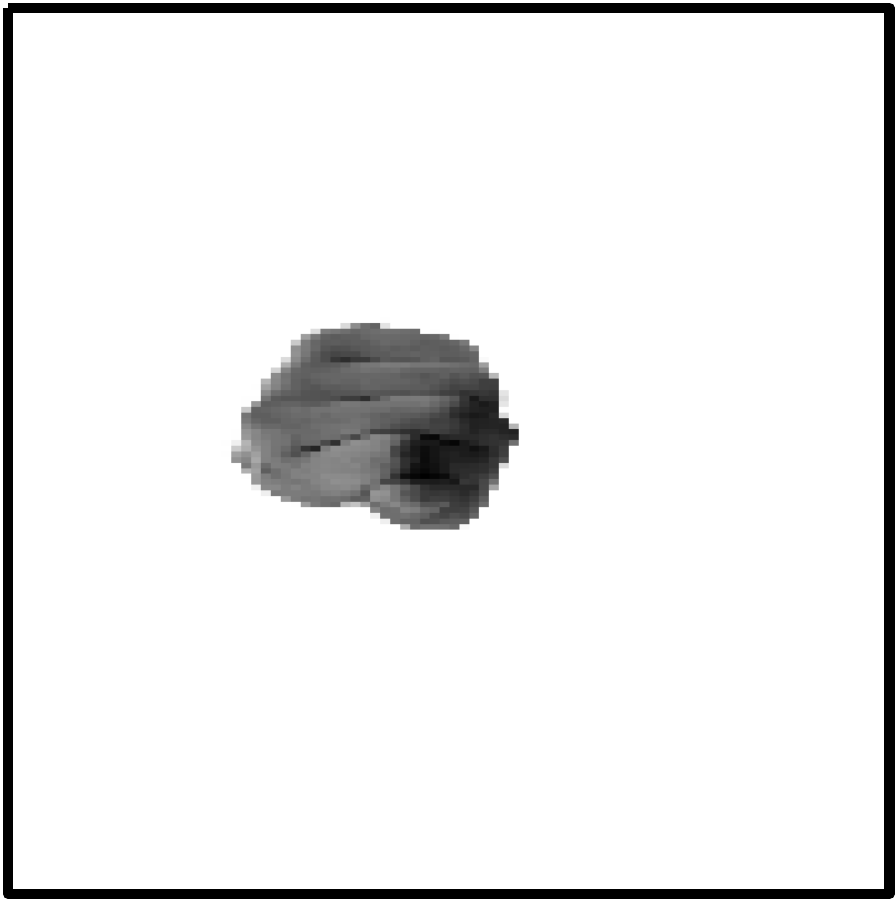}&
\includegraphics[width=0.19\textwidth]{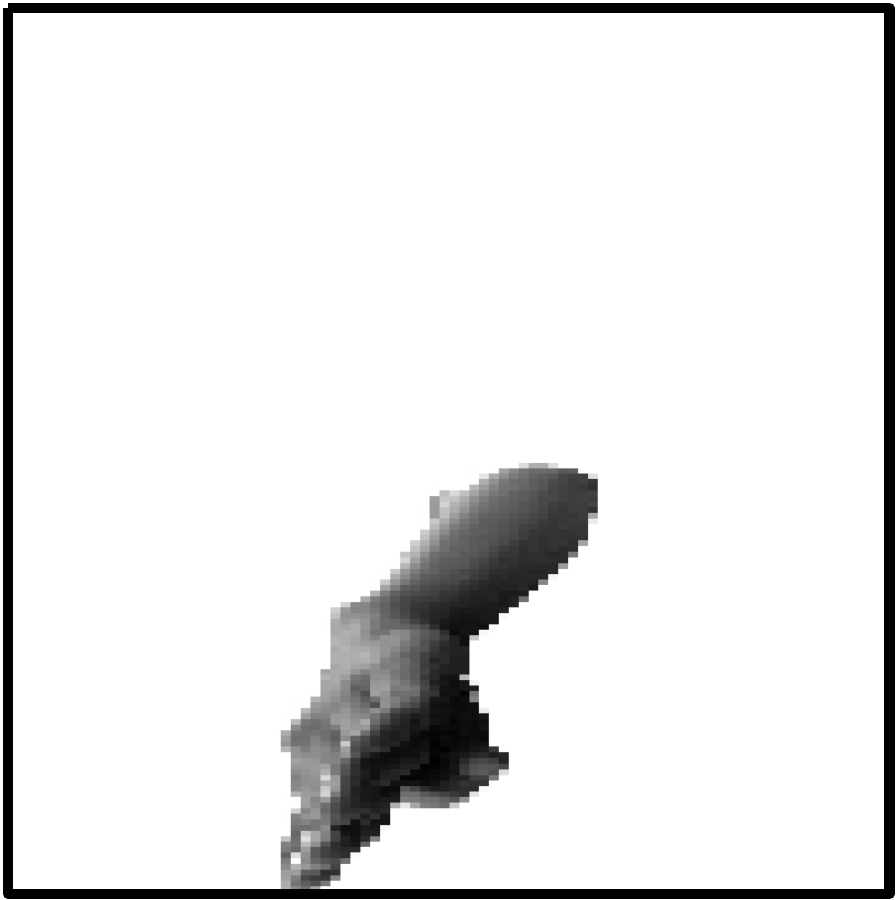}&
\includegraphics[width=0.19\textwidth]{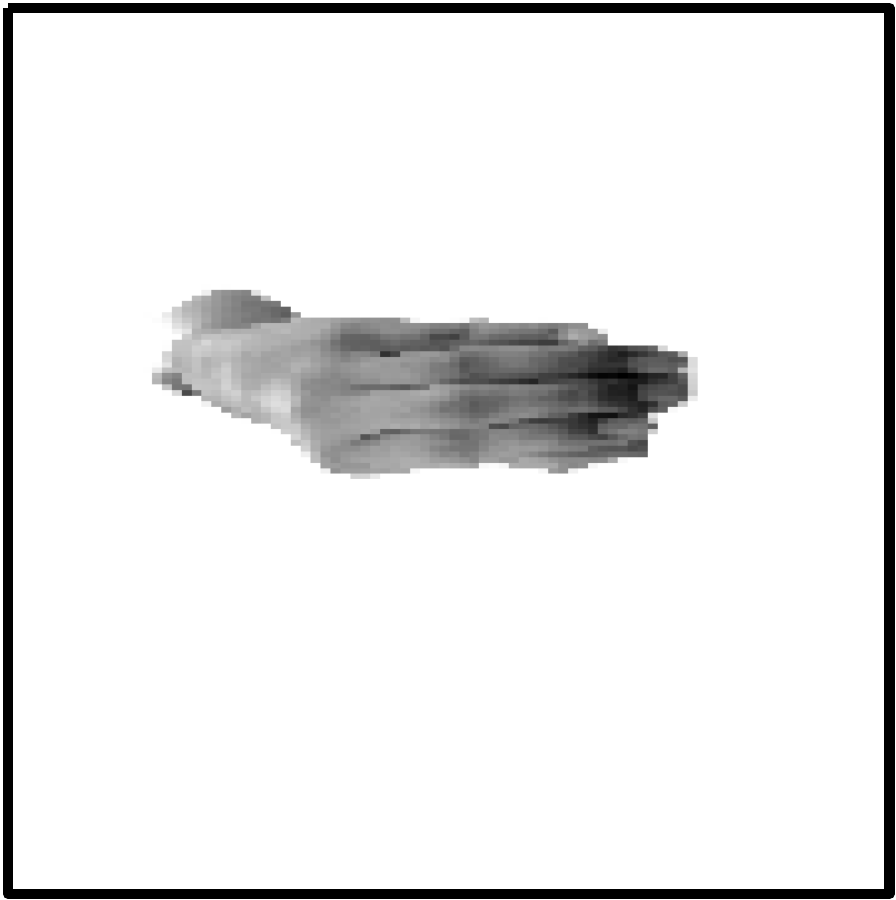}&
\includegraphics[width=0.19\textwidth]{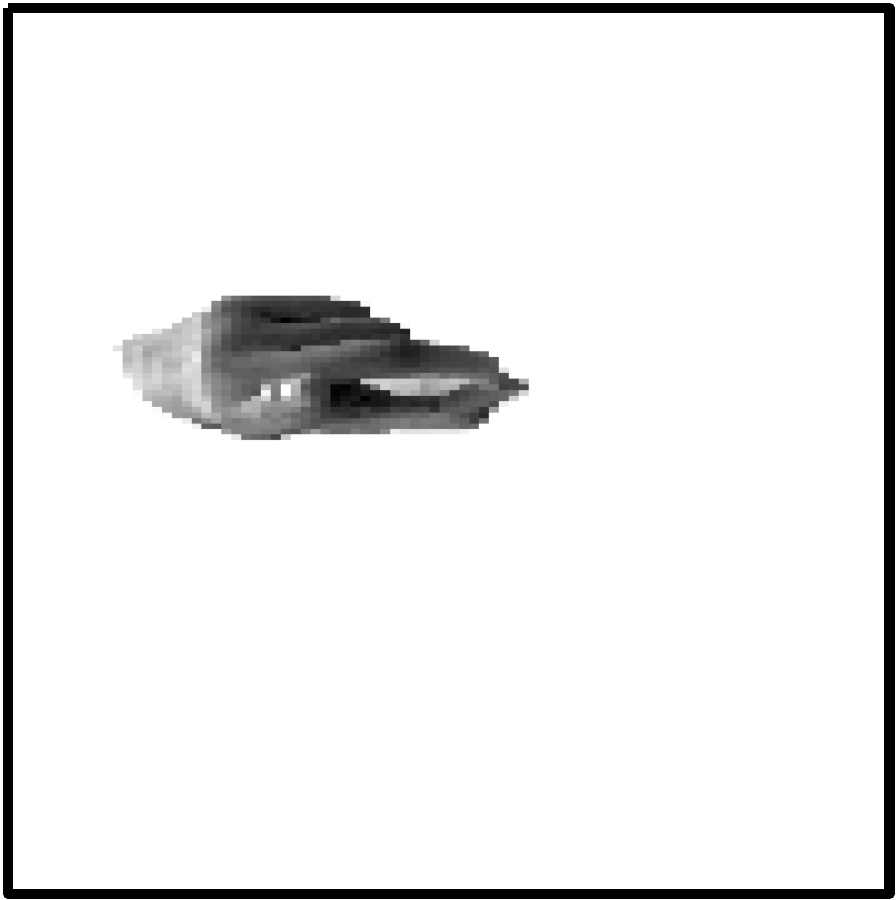}&
\includegraphics[width=0.19\textwidth]{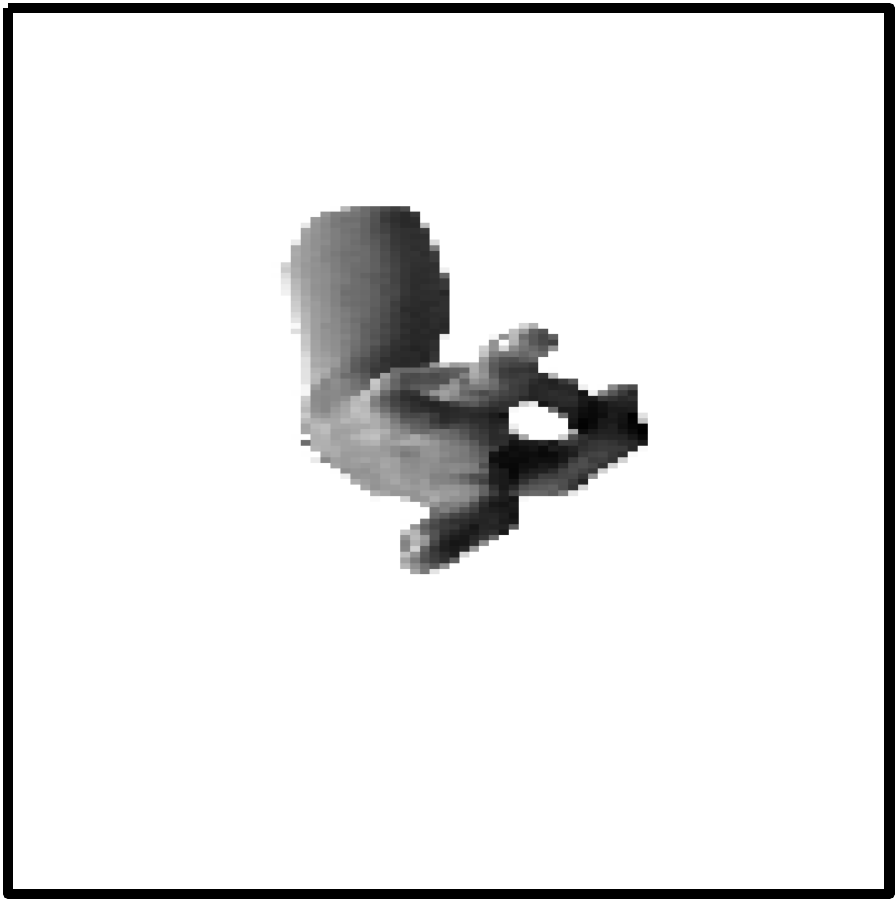}\\
(a) A & (b) B & (c) C & (d) D & (e) W
\end{tabular}
\vspace{-4pt}
\caption{Sample raw images in the gray-scale dictionary. The gestures from left to right correspond to the letters: A, B, C, D and W.}\label{fig3}
\end{figure*}

\begin{figure*}[h]
\centering\setlength{\tabcolsep}{2pt}
\begin{tabular}{ccccc}
\includegraphics[width=0.19\textwidth]{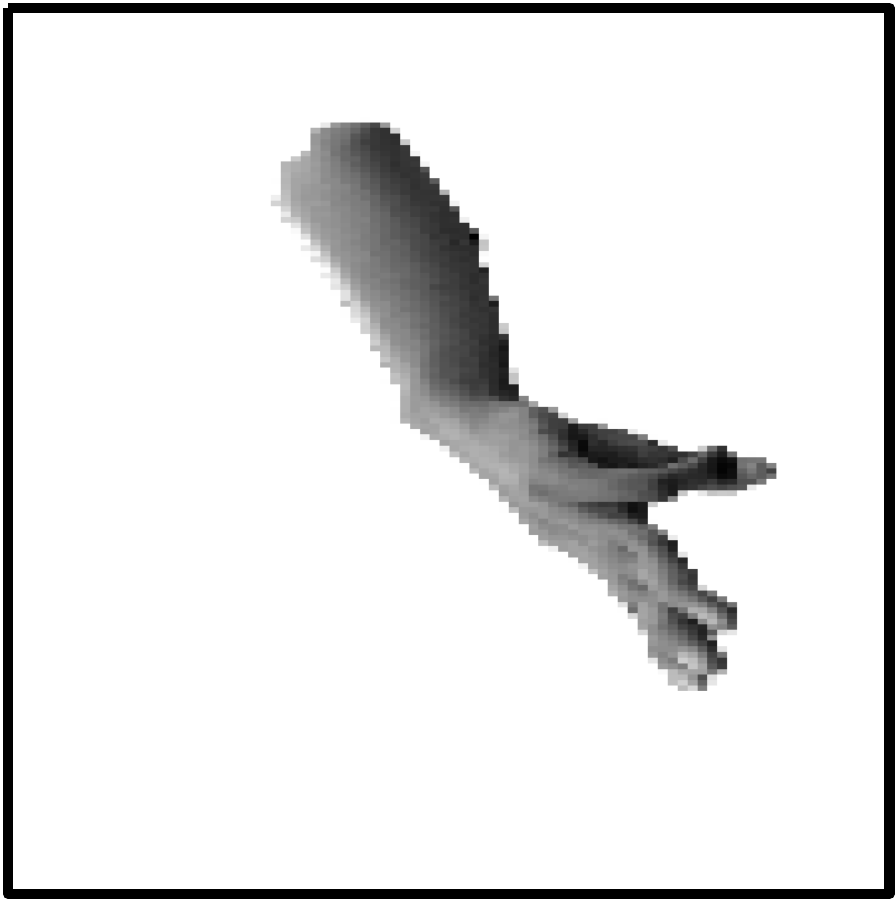}&
\includegraphics[width=0.19\textwidth]{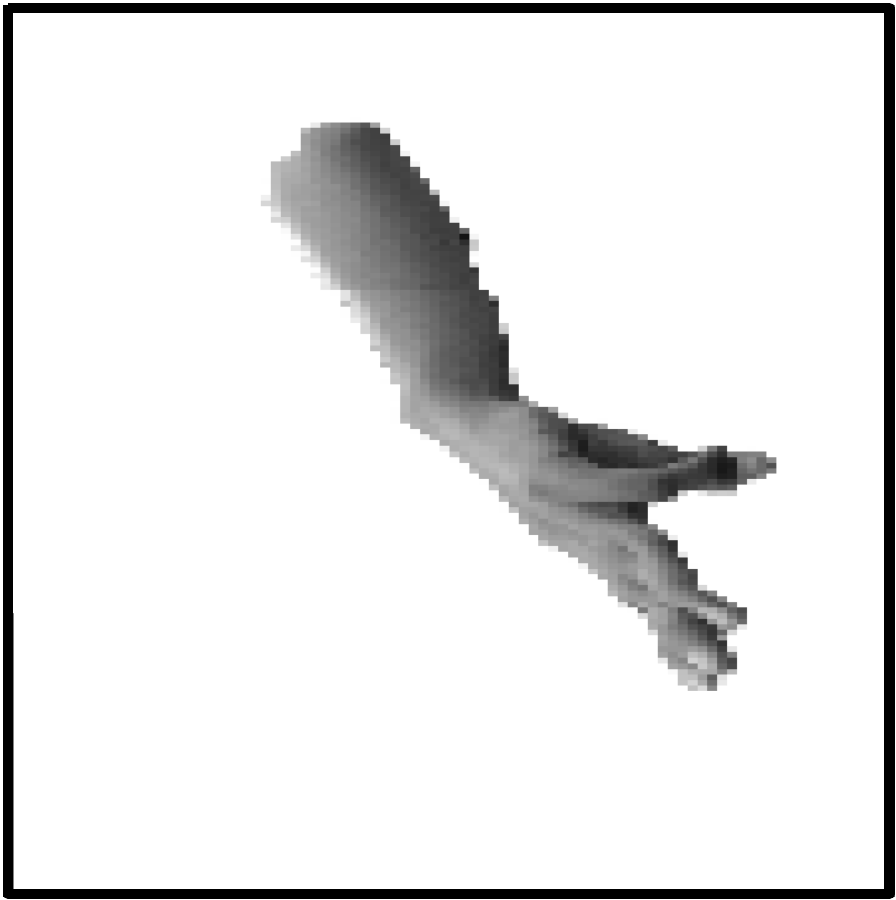}&
\includegraphics[width=0.19\textwidth]{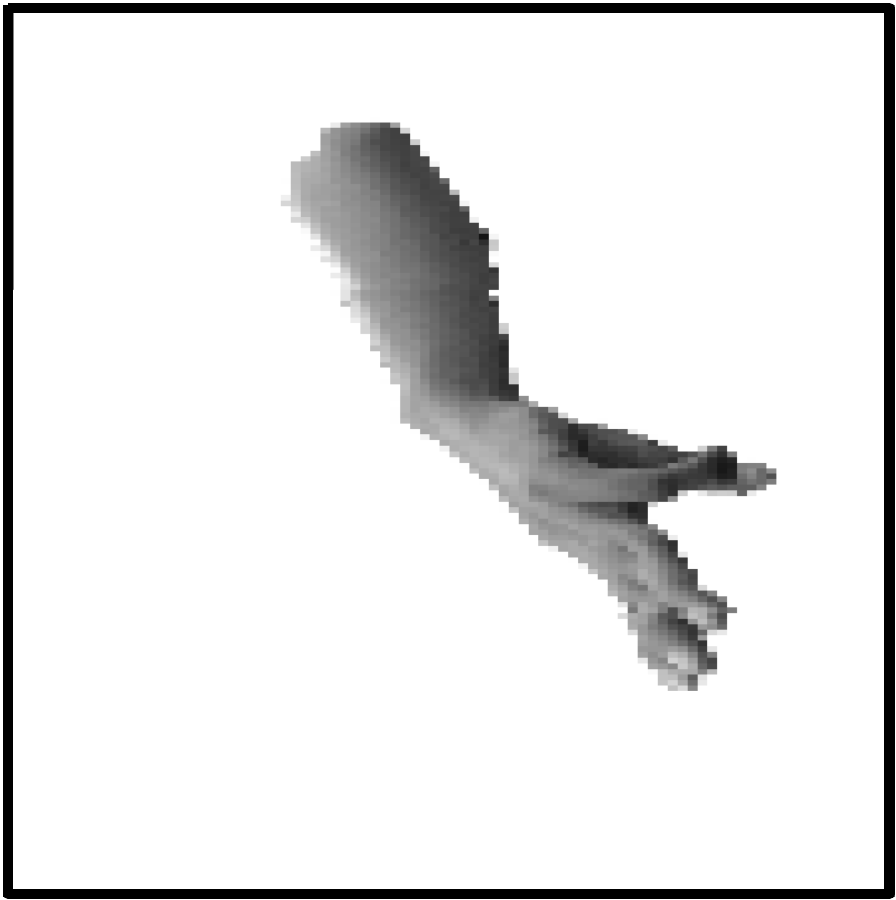}&
\includegraphics[width=0.19\textwidth]{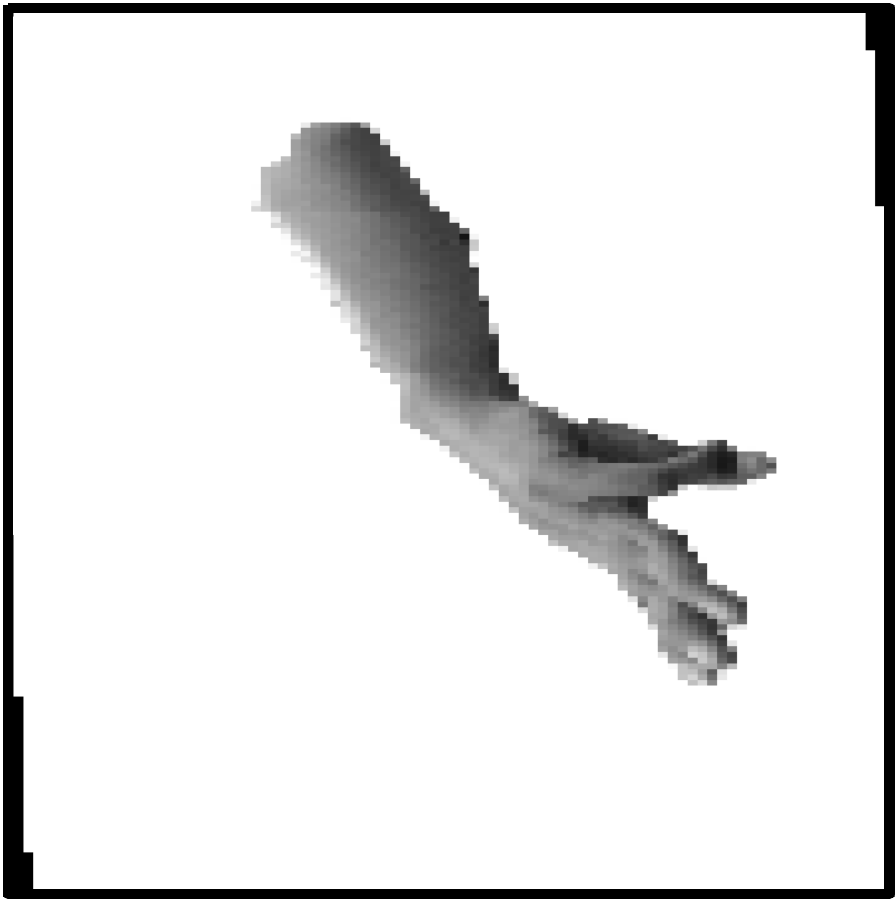}&
\includegraphics[width=0.19\textwidth]{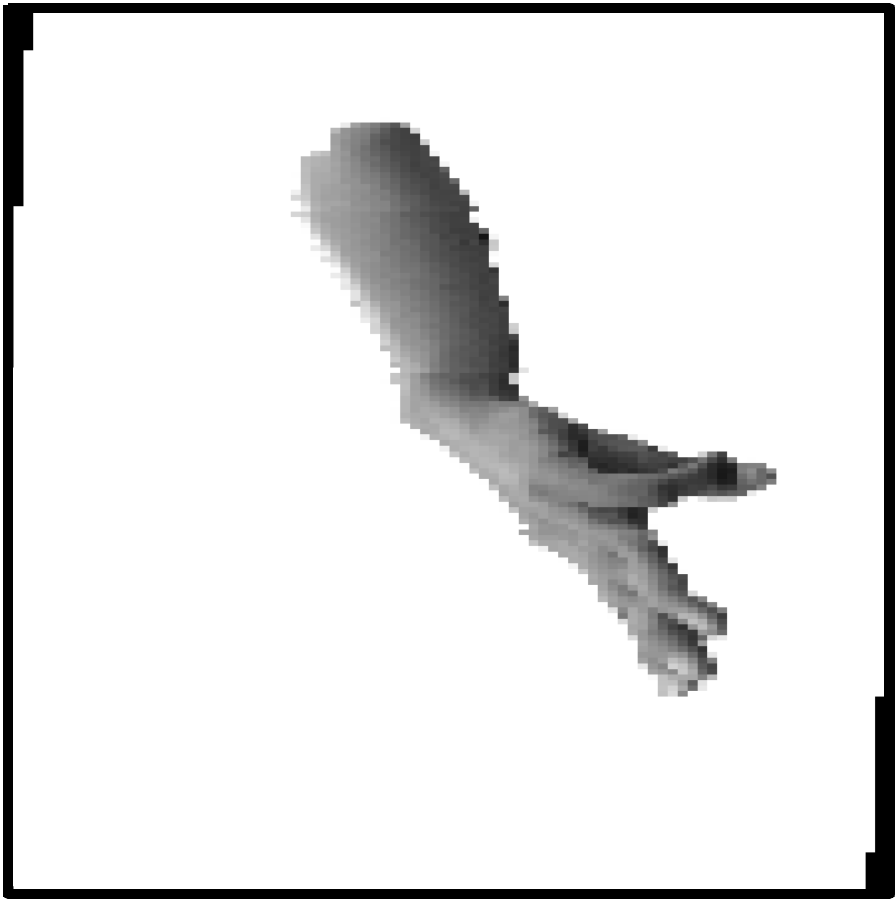}
\end{tabular}
\vspace{-4pt}
\caption{Image rotation. From left to right: original image, rotated by $1^\circ$, $-1^\circ$, $2^\circ$ and $-2^\circ$. Positive angles correspond to counterclockwise rotation and negative angles correspond to clockwise rotation.}\label{fig:rot}
\end{figure*}

\begin{figure}[H]
\centering
\includegraphics[width=.34\textwidth]{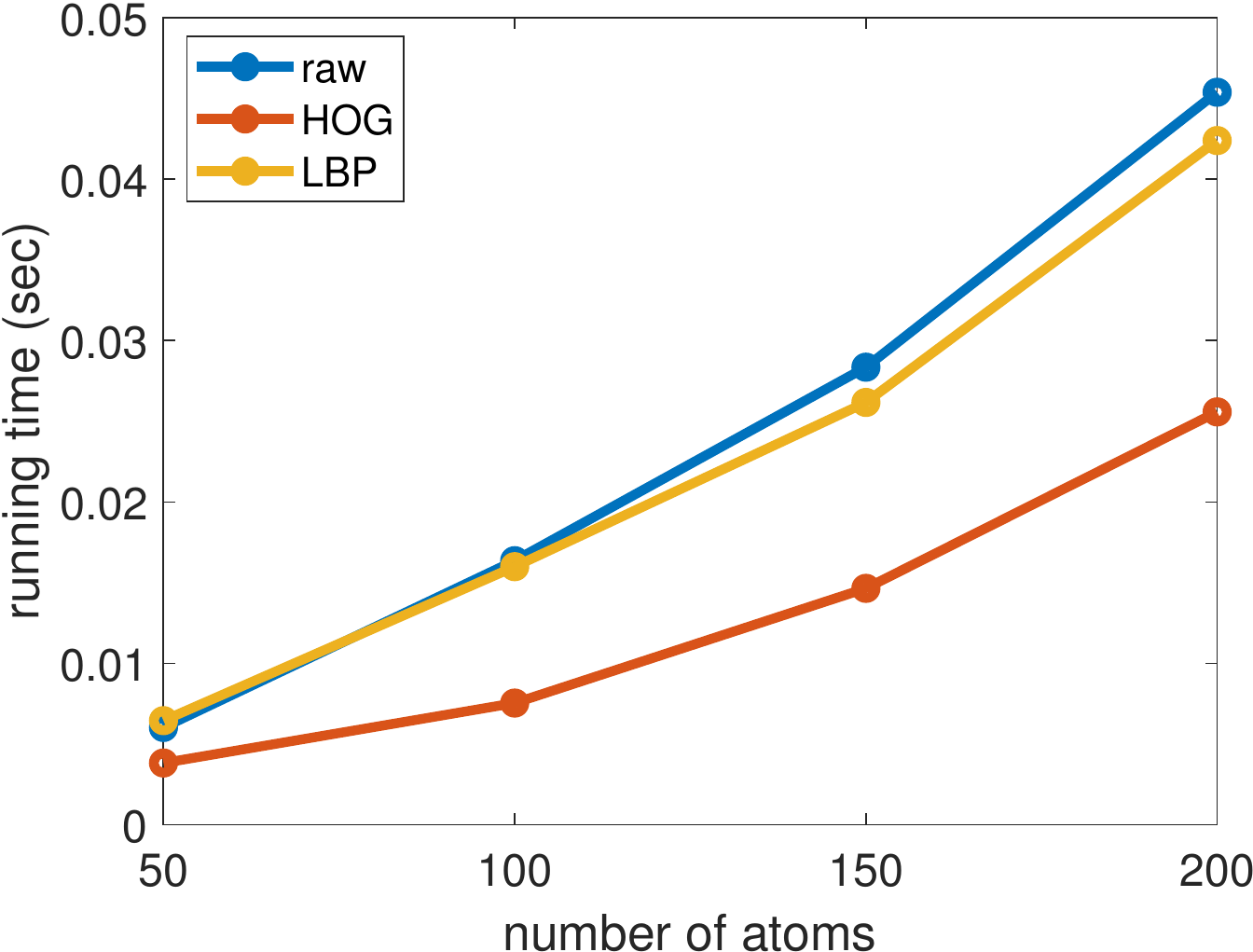}
\vspace{-8pt}
\caption{Running time comparison for a gray-scale dictionary.}\label{fig:exp2_time}
\vspace{-10pt}
\end{figure}

In the second experiment, we download the HGM-4 multi-cameras dataset \cite{hoang2020hgm} from \url{https://data.mendeley.com/datasets/jzy8zngkbg/1}. In particular, we choose five classes of images corresponding to the hand gestures that express the five letters: A, B, C, D and W. Each class of the original database has 40 atom images, each of size $160\times 90$. To ensure a sparse representation of atoms from the dictionary, we generate 67 additional images corresponding to those five gestures using a Logitech RGB webcam of resolution $1280\times 720$. A simple interface is developed to allow a user to classify and record gestures on their own using this webcam. All gestures are done with a whiteboard backdrop to reduce noise and help normalize the dataset. One set of such high-resolution images are shown in Fig.~\ref{fig2}. All newly generated images are resized to $160\times 90$. Thus far, we get a dataset with five classes, and the numbers of images within each class are distributed as 54, 52, 54, 54, 53. Furthermore, we expand the dataset by making four types of image rotations for each image in Matlab: clockwise/counterclockwise rotation by one/two degrees. Image rotation is illustrated in Fig.~\ref{fig:rot}. Note that rotation could bring zero boundary artifacts for large angles. Next we randomly select 10 test images from each class, and randomly select $Nt$ atom images from the rest of the class to form a test dictionary. We select $Nt=50,100,150,200$. A collection of five test gray-scale images are shown in Fig.~\ref{fig3}. In Table~\ref{tab2}, we list recognition rates for various numbers of atoms in each class of the dictionary using various types of features. One can see that HOG performs best most of the time and has a great advantage for small training sets. Raw feature in gray scale performs slightly better than LBP in this case due to the limited texture-like patterns. If the number of atoms is larger than 250, then all those three features yield almost perfect recognition. Running time for each test case is illustrated in Fig.~\ref{fig:exp2_time}. With the fixed cell size $8\times 8$, HOG feature vector has the smallest dimension among all the three feature types which indicates HOG takes the least running time and yields the best performance on average.

\begin{table}[h]
\centering
\caption{Recognition rates on a gray-scale dictionary.} \label{tab2}
\vspace{-6pt}
\begin{tabular}{ccccc}
\hline \hline
Feature $\backslash$ Atom No. & 50 & 100 & 150 & 200 \\ \hline
raw & 0.7124 & 0.9100 & 0.9756 & 0.9956 \\
HOG & 0.7360 & 0.9140 & 0.9832 & 0.9972 \\
LBP & 0.7140 & 0.8940 & 0.9664 & 0.9936 \\
\hline\hline
\end{tabular}
\vspace{-6pt}
\end{table}

\subsection{Discussions}
In this section, we discuss selection of parameters, cell size in HOG/LBP feature extraction and identification criteria. In addition, we make a remark about the comparison of $\ell_1$ and $\ell_{1-2}$ in our method.

\begin{figure}[H]
\centering
\includegraphics[width=.2\textwidth]{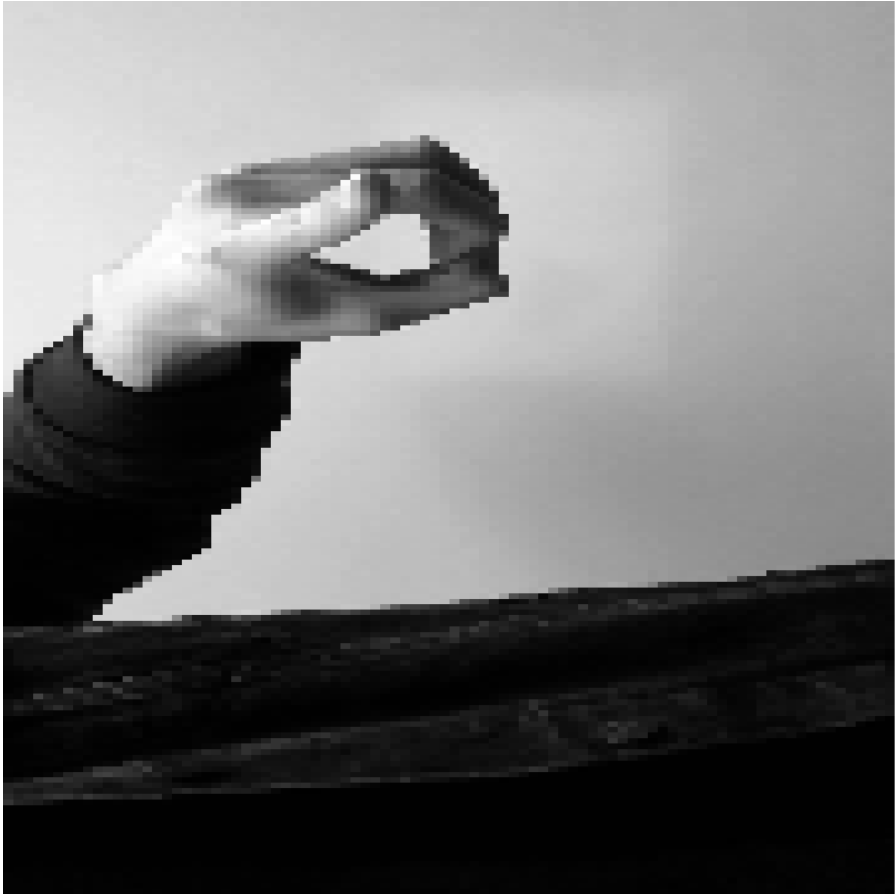}
\vspace{-6pt}
\caption{One example that is sensitive to parameter selection.}\label{fig:exp3_d1}
\vspace{-6pt}
\end{figure}

\paragraph*{Parameter Selection} In Algorithm~\ref{alg}, $\lambda$ is a regularization parameter which controls the balance between the data fidelity and the sparsity. The larger the parameter $\lambda$ is, higher sparsity is enforced to the desired vector $\vx$ but with larger residual error. In other words, if the test image is very similar to one atom in the test dictionary, then we could choose a large value for $\lambda$. In the $\vx$-update \eqref{eqn:x_update}, $\rho$ can be set as a large number to penalize the high sparsity so that the objective function decays fast. The number of inner loops could be set to be a small integer when it decays fast. Further, if the background is not removed, then the recognition could be sensitive to the parameter selection. As one illustrative example, Fig.~\ref{fig:exp3_d1} has the ground truth gesture ``D'' which can be mistakenly recognized as ``C'' with HOG/LBP features of cell size $16\times 16$ unless we choose the parameters $\lambda=\rho=1$ and $N_{in}=100$. Here we use the HGM-4 database as in Section~\ref{subsec:exp2} with 200 images in each class. In this case, we can either preprocess the test image by removing the background or tune parameters carefully.

\paragraph*{Cell Size in HOG and LBP} The cell size in HOG and LBP feature extraction influences the running time of Algorithm~\ref{alg} and its performance on parameter-sensitive test images, e.g., Fig.~\ref{fig:exp3_d1}. There is a trade-off between computational cost and recognition accuracy. Large cell sizes will yield low-dimensional features and thereby alleviate the computational burden, which however may cause inaccuracies in local description and reduce the recognition rate. Dimensions for various types of features in our experiments are listed in Table~\ref{tab3}. A cell of size $k\times k$ for both HOG and LBP features in the range 8$\sim$28 works in most situations. %In our Experiment~2, the cell size $k=20$ works most frequently for our generated test images.

\begin{table}[H]
\centering
\caption{Dimensions of Various Feature Vectors}\label{tab3}
\vspace{-6pt}
\begin{tabular}{ccccc}
\hline\hline
 $k$ & 8 & 12 & 16 & 20\\ \hline
\multicolumn{5}{c}{image size $150\times 150$}\\ \hline
HOG & 10404 & 4356 & 2304& 1296\\
LBP & 19116 & 8496 & 4779 &2891\\ \hline
\multicolumn{5}{c}{image size $160\times 90$}\\ \hline
HOG & 6840 & 2592 & 1296 & 756\\
LBP & 12980 & 5369 & 2950 & 1888\\
\hline\hline
\end{tabular}
\vspace{-6pt}
\end{table}

\paragraph*{Identification Metric} Two types of identification metrics are introduced in the paper, including $\ell_2$-norm based residual \eqref{eqn:ri1} and cosine similarity based metric \eqref{eqn:ri2}. According to our numerical experiments, these two metrics achieve almost the same recognition performance. However, it is worth noting that \eqref{eqn:ri1} may result in a very large number while \eqref{eqn:ri2} is always between 0 and 1. To avoid numerical instability issues, \eqref{eqn:ri2} could be a top choice.

\paragraph*{Comparison of $\ell_1$ and $\ell_{1-2}$ regularizations} The $\ell_{1-2}$-regularization $\norm{\cdot}_1-\beta\norm{\cdot}_2$ can be reduced to the $\ell_1$-regularization when $\beta=0$, and it is also related to the iterative reweighted $\ell_1$ (IRL1) \cite{candes2008enhancing,guo2021novel} by choosing a special weighting scheme. Our vast experiments have shown that $\ell_{1-2}$ regularization performs slight better than $\ell_1$ in the same algorithmic framework especially with LBP features. For instance, Table~\ref{tab4} shows the recognition rates for Algorithm~\ref{alg} with $\ell_1$ and $\ell_{1-2}$ regularizations and LBP features using the same data and parameter setting as in Section~\ref{subsec:exp2}. This phenomenon can be interpreted by the fact that both regularizations could lead to the solutions with same sparsity level, which will not significantly impact the recognition. Nevertheless, $\ell_{1-2}$ will converge to a local minimizer faster than $\ell_1$ and thus is more efficient when only a few training data is available.

\begin{table}[h]
\centering
\caption{Recognition rate comparison for $\ell_1$ and $\ell_{1-2}$}\label{tab4}
\vspace{-6pt}
\begin{tabular}{ccccc}\hline\hline
Atom No. & 50 & 100 & 150 & 200 \\ \hline
$\ell_1$ & 0.6620 & 0.8940 & 0.9640 & 0.9920\\
$\ell_{1-2}$ & 0.6640 & 0.8980 & 0.9640 & 0.9940\\ \hline\hline
\end{tabular}
\vspace{-6pt}
\end{table}

\section{Conclusions and Future Work}\label{sec:con}
Vision-based hand gesture recognition has been widely in a lot of human-robot interaction applications. When there are only a limited number of training samples available, it becomes challenging to accurately detect the class of a given hand gesture image. In this paper, we propose a novel hand gesture recognition approach based on the nonconvex $\ell_{1-2}$ regularization to improve the performance. Compared to the $\ell_1$-regularization, $\ell_{1-2}$-regularization in the form of the difference of two vector norms can further promote sparsity which can enhance the prediction accuracy and/or achieve fast convergence to a local minimizer. To solve the $\ell_{1-2}$-regularized recognition model, we apply the ADMM framework which leads to two subproblems at each iteration. One subproblem is a least-square problem that has a closed-form solution by solving its normal equation. The other is reduced to the proximal operator of the $\ell_{1-2}$ regularizer which can be expressed by the shrinkage operator. To make the proposed method robust, we consider three types of features, including raw images in either binary or gray scale, HOG and LBP features. Numerical experiments on two data sets with various settings have shown the proposed effectiveness. In the future work, we will explore hybrid types of features by concatenating multiple features such as fusion of HOG and LBP, compare $\ell_1$ and $\ell_{1-2}$ regularizations in more settings, and extend the proposed framework to solve other related recognition problems, e.g., arm gesture recognition.

\addtolength{\textheight}{-12cm}   % This command serves to balance the column lengths
                                  % on the last page of the document manually. It shortens
                                  % the textheight of the last page by a suitable amount.
                                  % This command does not take effect until the next page
                                  % so it should come on the page before the last. Make
                                  % sure that you do not shorten the textheight too much.

%%%%%%%%%%%%%%%%%%%%%%%%%%%%%%%%%%%%%%%%%%%%%%%%%%%%%%%%%%%%%%%%%%%%%%%%%%%%%%%%

%%%%%%%%%%%%%%%%%%%%%%%%%%%%%%%%%%%%%%%%%%%%%%%%%%%%%%%%%%%%%%%%%%%%%%%%%%%%%%%%

%%%%%%%%%%%%%%%%%%%%%%%%%%%%%%%%%%%%%%%%%%%%%%%%%%%%%%%%%%%%%%%%%%%%%%%%%%%%%%%%
%\section*{APPENDIX}
%
%Appendixes should appear before the acknowledgment.

\section*{ACKNOWLEDGMENTS}

The research of Qin is supported by the NSF grant DMS-1941197 and the research of Ashley and Xie is supported by Woodrow W. Everett, Jr. SCEEE Development Fund in cooperation with the Southeastern Association of Electrical Engineering Department Heads.

%%%%%%%%%%%%%%%%%%%%%%%%%%%%%%%%%%%%%%%%%%%%%%%%%%%%%%%%%%%%%%%%%%%%%%%%%%%%%%%%

\bibliographystyle{IEEEtran}
\bibliography{reference}

\end{document}